\documentclass[runningheads]{llncs}

\usepackage{eccv}

\usepackage{eccvabbrv}

\usepackage{graphicx}
\usepackage{booktabs}

\usepackage[accsupp]{axessibility}  

\usepackage{colortbl}
\usepackage{graphicx}
\usepackage{soul}
\usepackage{enumitem}
\usepackage{bm}
\usepackage{lipsum}
\usepackage{multirow}
\usepackage{amsmath}
\usepackage{bbm}
\usepackage{pifont} 
\usepackage{caption}
\usepackage{marvosym}
\usepackage{fontawesome5}
\usepackage{pifont}
\usepackage{anyfontsize} 

\usepackage{hyperref}

\usepackage{orcidlink}

\begin{document}

\title{C2E: Boosting Ego-Only 3D Object Detection via Multi-Teacher Contrastive Knowledge Distillation} 

\titlerunning{C2E}

\author{Jinlong Wang\inst{1,2}\textsuperscript{\(\star\)}\orcidlink{0009-0005-1074-2854} \and
Xun Huang\inst{1,2,3}\textsuperscript{\(\star\)}\orcidlink{0009-0007-5343-7811} \and 
Qiming Xia\inst{1,2}\orcidlink{0000-0002-6440-5694} \and \\ 
Shijia Zhao\inst{1,2}\orcidlink{0000-0002-1132-134X} \and Chenglu Wen\inst{1,2}\textsuperscript{\Letter}\orcidlink{0000-0002-6189-1236} }   

\authorrunning{J.~Wang et al.}

\institute{Fujian Key Laboratory of Urban Intelligent Sensing and Computing, \\ Xiamen University, China \and
Key Laboratory of Multimedia Trusted Perception and Efficient Computing, \\Ministry of Education of China, Xiamen University, China \\ \and
Zhongguancun Academy, China\\}

\maketitle
\renewcommand{\thefootnote}{} 
\footnotetext[1]{$^{\star}$~Equal contribution.}
\footnotetext[1]{\Letter~Corresponding author: \email{clwen@xmu.edu.cn}.}


\begin{abstract}
LiDAR-based 3D object detection is essential for autonomous driving systems. However, traditional Ego-only Perception (Eo-Perception) suffers from limited perspective and occlusions in a complex outdoor environment, leading to performance bottlenecks. Recently, research on multi-agent Collaborative Perception (Co-Perception) has demonstrated excellent performance, but high communication costs and accumulated pose error hinder its application. To address this, we explore a novel \textbf{C2E} (\textbf{C}o-Perception \textbf{to} \textbf{E}o-Perception) paradigm through the \textbf{M}ulti-\textbf{to}-\textbf{S}ingle (\textbf{M2S}) agent contrastive knowledge distillation framework. Our M2S framework first designs Multi-Level Feature Enhancement module to provide more stable features, and introduces Auxiliary Point Cloud Reconstruction and Multi-Teacher Contrastive Distillation mechanisms to mitigate domain gaps in point cloud and feature distributions within the C2E paradigm. Benefiting from this, our M2S can retain the excellent performance of collaborative perception while effectively avoiding the drawbacks, such as communication delays and positioning errors. Extensive experiments on the V2XSet, V2V4Real and DAIR-V2X datasets show the effectiveness and generalizability of our M2S framework when combined with the state-of-the-art CoSDH model and other excellent 3D detectors. Our M2S framework can deliver up to a 8.64\% improvement in 3D mAP performance without introducing any communication costs. 

\end{abstract}    
\section{Introduction}

3D object detection is a crucial task for autonomous driving and other unmanned systems~\cite{survey_3DOD_robust_lidar,L4dr,Towards2024}. Recently, LiDAR-based methods have achieved remarkable progress~\cite{survey_ijcv23, hinted} and have been widely adopted, owing to the precise spatial information provided by LiDAR sensors~\cite{Lidar_for_autonomous_driving}.

Outdoor environments, however, present complex and dynamic challenges, including diverse occlusions and extreme long-range detection requirements. These factors significantly impact Ego-only Perception (\textcolor{cyan}{Eo}\textcolor{orange}{-}\textcolor{cyan}{P}erception), leading to inaccurate perception results ~\cite{occluded_objects_detection,survey_co_IV}. While previous research~\cite{boost_frame, virconv, boost_modal, 1-supfusion, 2-2dpass } has attempted to alleviate these challenges of Eo-Perception by using more information or multi-sensor distillation, a noticeable performance bottleneck remains.

\begin{figure}[!t]
    \includegraphics[width=1.0\columnwidth]{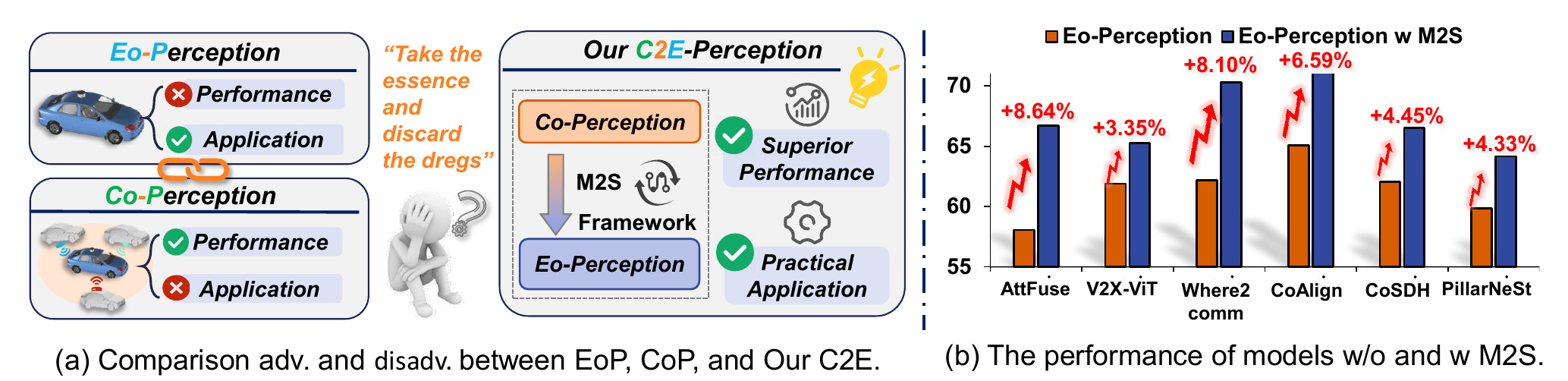}
  \caption{\textcolor{black}{(a) Comparison of the advantages and limitations between Eo-perception, Co-perception, and our proposed C2E-perception. (b)  The 3D mAP$@$0.7 performance of SOTA models without and with our M2S framework on the V2XSet. } }
  \label{motivation}
\end{figure}

Recently, some efforts have been made to investigate Collaborative Perception (\textcolor{ForestGreen}{Co}\textcolor{orange}{-}\textcolor{ForestGreen}{P}erception), such as vehicle-to-vehicle (V2V), vehicle-to-infrastructure (V2I), and vehicle-to-everything (V2X) ~\cite{V2X-ViT,attfuse,V2X-R_CVPR, Dota}. Leveraging shared information among agents has demonstrated outstanding performance in the domain of cooperative 3D object detection. Nonetheless, the collaborative challenges that arise in real-world scenarios, such as high communication costs and localization errors, continue to hinder its real-world applications significantly.

Consequently, several existing methods have attempted to address the collaboration challenges by studying how, when, and where to communicate or by compressing communication features ~\cite{yang2023how2comm, yang2023what2comm, where2comm, bm2cp, coalign}. Nevertheless, despite some progress, these collaboration challenges still hinder the practical application of collaborative 3D object detection, primarily due to the restrictive nature of the existing collaboration mechanisms~\cite{survey_CO_Methods_datasets_challenges}.

Based on existing research, we have summarized the strengths and weaknesses of these two perception paradigms. As illustrated in Fig. \ref{motivation}(a), Eo-Perception is practical but has performance bottlenecks, whereas Co-Perception delivers superior performance while encountering collaboration challenges. "\textit{Take the essence and discard the dregs.}" Can the advantages of both be combined? This is the core question this paper addresses. To this end, we explore a novel perception paradigm, \textbf{\textcolor{ForestGreen}{C}\textcolor{orange}{2}\textcolor{cyan}{E}-Perception}: transferring the superior performance of Co-Perception to practical Eo-Perception. Specifically, at the training stage, we retain high-density information and collaborative knowledge from the Co-Perception teacher to the Eo-Perception student through the knowledge distillation technique. At the inference stage, we use the Eo-Perception student for individual inference to mitigate collaboration challenges.

However, simple knowledge distillation faces substantial domain gaps in data and feature distributions between single-agent and multi-agent data, which limits the effective transfer of collaborative perception knowledge. To bridge these gaps, we introduce the Auxiliary Point Cloud Reconstruction (APCR)  and  Multi-Teacher Contrastive Distillation (MTCD) strategies to mitigate the significant discrepancy in data and feature distributions, respectively. Moreover, to provide more stable features for distillation, we propose Multi-Level Feature Enhancement (MLFE) to enhance the feature extraction capability of the student. 

As shown in Fig. \ref{motivation}(b), extensive experiments conducted on both real-world and simulated datasets demonstrate the effectiveness and generalizability of our M2S framework when integrated with the state-of-the-art model ~\cite{cosdh} and other excellent 3D detectors. 
In summary, our main contributions are as follows.

\begin{itemize}[leftmargin=10pt]

\item To the best of our knowledge, we are the first to explore the Co-Perception to Eo-Perception (\textbf{C2E-Perception}) paradigm. We propose the novel M2S framework based on multi-teacher adaptive contrastive knowledge distillation.
\item We effectively mitigate the significant data and feature gaps between multi-agent and single-agent domain through Auxiliary Point Cloud Reconstruction (APCR) and Multi-Teacher Contrastive Distillation (MTCD). Moreover, we use the Multi-Level Feature Enhancement (MLFE) to provide more stable features for distillation.
\item Extensive experiments on the V2XSet, V2V4Real, and DAIR-V2X datasets validate the effectiveness of the proposed M2S framework. Our M2S framework can deliver an improvement of up to 8.64 in 3D mAP performance without introducing any communication costs.

\end{itemize}
\section{Related work}
\textbf{3D Object Detection for Single Agent.}
3D object detection aims to identify object categories and spatial locations from sparse and irregular point clouds. Existing single-agent methods fall into three categories: point-based~\cite{pointrcnn,point-gnn}, grid-based~\cite{second,CenterPoint,L4dr,coin}, and point-voxel hybrid approaches~\cite{pvrcnn,voxelrcnn}. However, single-agent systems rely solely on onboard sensors, resulting in limited perception angles and susceptibility to occlusions, blind spots, and complex traffic dynamics~\cite{survey_co_IV}. Furthermore, the limited sensing range constrains distant object detection~\cite{survey_CO_Methods_datasets_challenges}, causing performance bottlenecks in high-speed or sparse scenes.

\noindent \textbf{Collaborative Perception for Multi-Agents.}
Collaborative perception has been proposed to overcome the limitations of a single viewpoint. Depending on the fusion stage, it can be categorized into early~\cite{disconet,cooper,Double-M_Quantification}, intermediate~\cite{attfuse,cosdh,V2X-R_CVPR,adafusion,V2X-ViT}, and late fusion~\cite{mash,mamp}. Despite its advantages, collaborative perception faces challenges such as high bandwidth demands for feature or raw data transmission~\cite{survey_Towards_V2X} and accumulated errors from inaccurate relative pose estimation~\cite{coalign,survey_CO_Methods_datasets_challenges}. DiscoNet~\cite{disconet} reduces communication costs via a distilled graph; V2X-INCOP~\cite{5-V2X-INCOP} improves fault tolerance under interruptions, yet both remain in the collaborative paradigm with latency and infrastructure constraints. Thus, improving perception performance while reducing communication overhead remains a key research focus.

\noindent \textbf{3D Object Detection Based on Knowledge Distillation.}
Knowledge distillation has been applied to 3D object detection~\cite{boost_frame,boost_modal,SRKD} and recent collaborative perception~\cite{disconet,Di-v2x}.
Multi-teacher schemes yield richer knowledge than single-teacher ones; MKD-Cooper~\cite{MKD-cooper} exploits this for robustness.
Unlike~\cite{3-learning}, which refines other agents' predictions as pseudo-labels, our method distills collaborative knowledge directly into the ego model. We thus propose M2S, a multi-teacher adaptive contrastive distillation framework that transfers dense knowledge from multiple teachers to a sparse student.
\begin{figure*}[!t]
    \includegraphics[width=\textwidth]{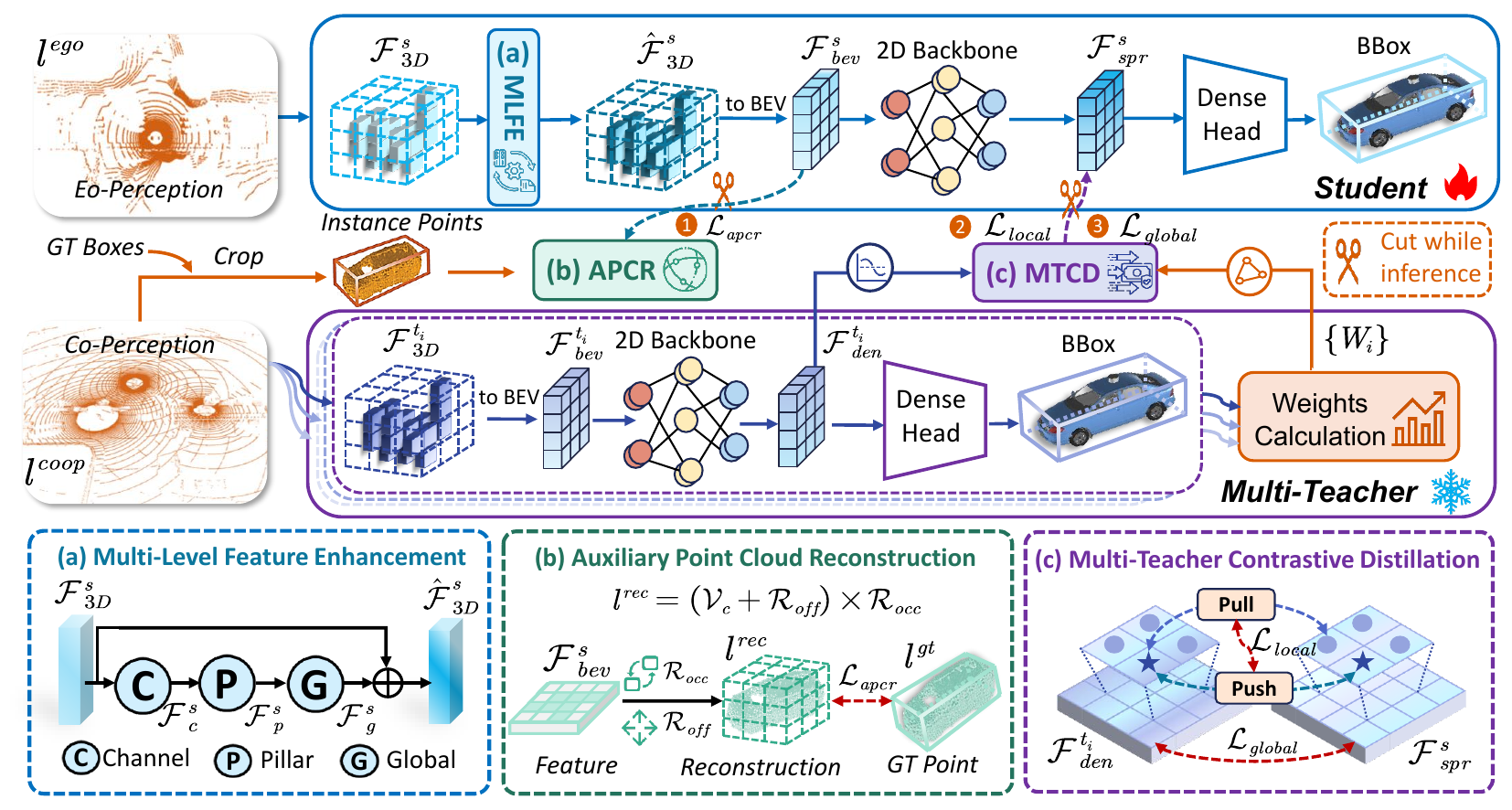}
  \centering
  \caption{\textbf{M2S framework}. 
  \textbf{Teachers} (Bottom): Multi-agent point clouds $ l^{coop} $ are simultaneously input into three teachers, which are then compressed into BEV features $ \{\mathcal{F}_{bev}^{t_i}\}_{i=1}^3 $ via pillar transformation. Then a 2D backbone is used to extract dense features $ \{\mathcal{F}_{den}^{t_i}\}_{i=1}^3 $; Finally, the MTCD module fuses them into a distilled supervision feature map $ \hat{\mathcal{F}}_{den}^{t} $. 
  \textbf{Student} (Top): The ego-only point cloud $l^{ego}$ is converted into pillars $ \mathcal{F}_{3D}^{s} $ and enhanced by the MLFE to obtain $ \hat{\mathcal{F}}_{3D}^{s}  $. During training, $ \hat{\mathcal{F}}_{3D}^{s} $ is projected to BEV space for voxel-level reconstruction via APCR, and the resulting features $ \mathcal{F}_{spr}^{s} $ are guided by $ \hat{\mathcal{F}}_{den}^{t} $ through multi-teacher adaptive contrastive distillation.
  }
  \label{framework}
\end{figure*}
\section{Methodology}

\subsection{Problem Statement and Overall Design}
\textbf{Problem Statement.}
In collaborative perception based on V2X communication, there are three types of agents: Ego, Connected Automated Vehicles (CAVs), and Infrastructure units (Inf). In our proposed framework, the collaborative perception (Co-Perception) teacher detector $\mathcal{D}^{t}$ receives lidar point cloud input $l^{coop} = \{l^{ego}, l^{cav}, l^{Inf}\}$ from all agents. In contrast, the ego-only perception (Eo-Perception) student detector $\mathcal{D}^{s}$ only takes point cloud data $\{l^{ego}\}$ from the ego vehicle, thus enabling single-agent 3D object detection. The goal of 3D object detection is to regress a set of 3D bounding boxes $\mathcal{B} = \{b_i\}_{i=1}^{N_b}$, where $\mathcal{B} \in \mathbb{R}^{N_b \times 7}$ and $N_b$ denotes the number of predicted bounding boxes. 

\textbf{Overall Design.}
Due to the varying number of participating agents, large gaps exist between $\mathcal{D}^{t}$ and $\mathcal{D}^{s}$ in both feature representation and point cloud domains. To address this challenge, we propose a \textbf{M}ulti-to-\textbf{S}ingle (M2S) agent contrastive knowledge distillation framework, as illustrated in Fig.~\ref{framework}.
First, the \textbf{M}ulti-\textbf{L}evel \textbf{F}eature \textbf{E}nhancement (MLFE) module (Fig.~\ref{framework}(a)) progressively enhances 3D features in $\mathcal{D}^{s}$ across multiple levels and dimensions.
Then, the \textbf{A}uxiliary \textbf{P}oint \textbf{C}loud \textbf{R}econstruction (APCR) module (Fig.~\ref{framework}(b)) encourages the student to learn point cloud distributions consistent with $\mathcal{D}^{t}$.
Finally, the \textbf{M}ulti-\textbf{T}eacher  \textbf{C}ontrastive \textbf{D}istillation (MTCD) strategy (Fig.~\ref{framework}(c)) transfers locally structured and globally salient features from the teacher model to the student via contrastive learning and adaptive weighting, which further mitigates the feature disparity between $\mathcal{D}^{t}$ and $\mathcal{D}^{s}$.
Note that only the student detector $\mathcal{D}^{s}$ is used during inference.

We provide the formal definitions of the three paradigms in the \textit{\textcolor{magenta}{Appx.}} and the following sections provide detailed descriptions of each module.

\subsection{Multi-Level Feature Enhancement} 
In the teacher $\mathcal{D}^{t}$, the ego vehicle obtains high-density features through communication with other agents. In contrast, the student $\mathcal{D}^{s}$ relies solely on its own sensors, producing relatively sparse features that limit 3D detection performance. To enhance the feature extraction capability of $\mathcal{D}^{s}$ and provide stable features for distillation, we propose the \textbf{M}ulti-\textbf{L}evel \textbf{F}eature \textbf{E}nhancement (MLFE) module, which progressively refines 3D features $\mathcal{F}_{3D}^{s}$ from local to global levels.  

\begin{figure}[!t]
\centering
    \includegraphics[width=1.0\columnwidth]{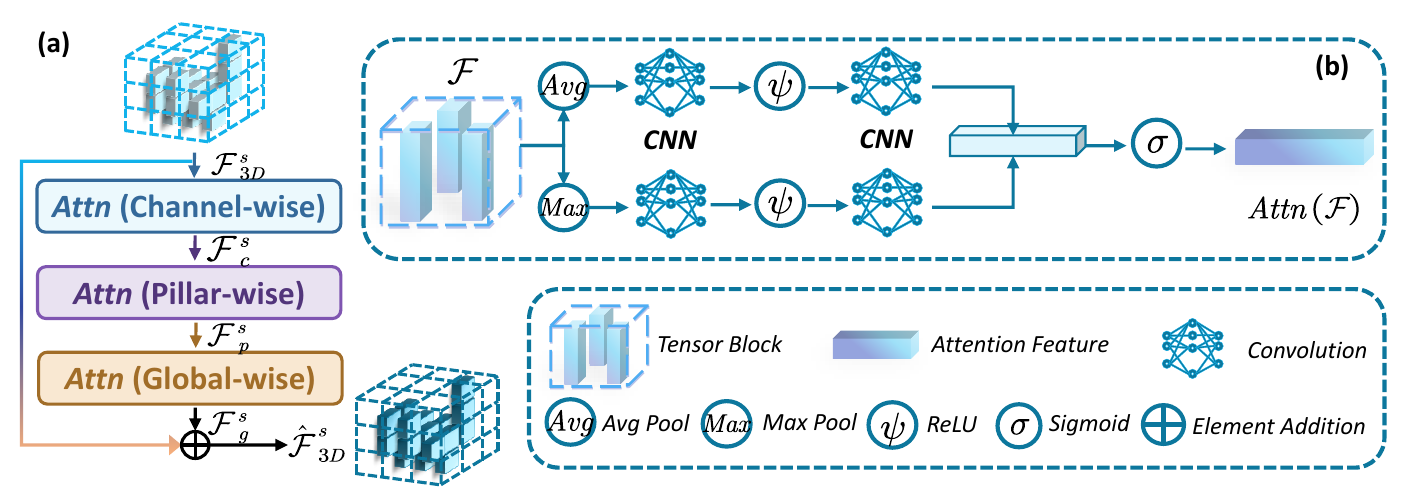}
  \caption{Multi-Level Feature Enhancement. (a) Serial enhancement strategy. (b) Feature weight calculation ($Attn$). We apply a serial local-to-global enhancement to the input $\mathcal{F}_{3D}^{s}$ in three stages: channel-wise, pillar-wise, and global-wise. }
  \label{mlfe}
\end{figure}

As shown in Fig.~\ref{mlfe}, MLFE enhances the student's 3D features along three dimensions: \textit{Channel}, \textit{Pillar}, and \textit{Global}, through hierarchical attention mechanisms. The student detector $\mathcal{D}^{s}$ receives a 3D feature tensor $\mathcal{F}_{3D}^{s} \in \mathbb{R}^{C \times P \times G}$, where $C$, $P$ and $G$ denote channels, points per pillar, and the numbers of global pillars, respectively. 
By progressively re-weighing semantic channels, reinforcing local geometric consistency, and integrating global contextual information, MLFE enhances students' representational capabilities, thereby generating more robust and domain-invariant features for distillation.

Inspired by SCNet3D~\cite{Scnet3d} and CBAM~\cite{cbam}, MLFE applies feature weight calculation ($Attn$) to generate three attention maps:
\begin{equation}
\begin{split}
Attn(\mathcal{F})
    &= \sigma ( \delta ( \psi \left( \delta \left(\theta_{max}(\mathcal{F}) \right) \right) \\
    & + \delta \left( \psi\left( \delta \left( \theta_{avg}(\mathcal{F}) \right) \right) \right),
\end{split}
\end{equation}
where $\theta_{max}$ and $\theta_{avg}$ denote max\_pooling and average\_pooling, $\delta$ denotes convolution, $\psi$ denotes ReLU activation, and $\sigma$ denotes sigmoid activation.

Then, we apply $Attn$ to $\mathcal{F}_{3D}^{s}$ along the channel, pillar, and global dimensions sequentially to  compute the attention map and refine features step-by-step as:

\begin{equation}
\begin{aligned}
\mathcal{F}_{c}^{s}=\mathcal{F}_{3D}^{s} \otimes Attn(\mathcal{F}_{3D}^{s}) \oplus \mathcal{F}_{3D}^{s},  
\\\mathcal{F}_{p}^{s}=\mathcal{F}_{c}^{s} \otimes Attn(\mathcal{F}_{c}^{s}) \oplus \mathcal{F}_{c}^{s} ,
\\\mathcal{F}_{g}^{s}=\mathcal{F}_{p}^{s} \otimes Attn(\mathcal{F}_{p}^{s}) \oplus \mathcal{F}_{p}^{s},
\\ \enspace\mathrm{and } \enspace \hat{\mathcal{F}}_{3D}^{s} =\mathcal{F}_{3D}^{s} \oplus \mathcal{F}_{g}^{s},
\end{aligned}
\end{equation}
where $ \otimes $ indicates element-wise multiplication and $ \oplus $ denotes element-wise addition. The final output $ \hat{\mathcal{F}}_{3D}^{s}  $ integrates attention-weighted 3D feature along all dimensions.

\subsection{Auxiliary Point Cloud Reconstruction} 
The teacher detectors $\mathcal{D}^{t}$ receives dense point cloud input $l^{coop}$ from multi-agent cooperation, while the student $\mathcal{D}^{s}$ only processes a sparse ego point cloud $l^{ego}$. This leads to a clear disparity in the \textbf{point cloud distribution domains} between the two networks. To address this gap, we introduce an \textbf{A}uxiliary \textbf{P}oint \textbf{C}loud \textbf{R}econstruction module (\textbf{APCR}) for the student model during the training phase. Unlike prior works~\cite{Sparse2Dense,2025aaaidsrc,r2ldm}, which focus on scene-level reconstruction, our method performs instance-level reconstruction between teacher and student. 
Specifically, we extract point clouds within ground-truth bounding boxes and re-voxelize them as supervision targets.

\begin{figure}[!t]
\centering
\begin{minipage}{0.45\columnwidth}
    \centering
    \includegraphics[width=\linewidth]{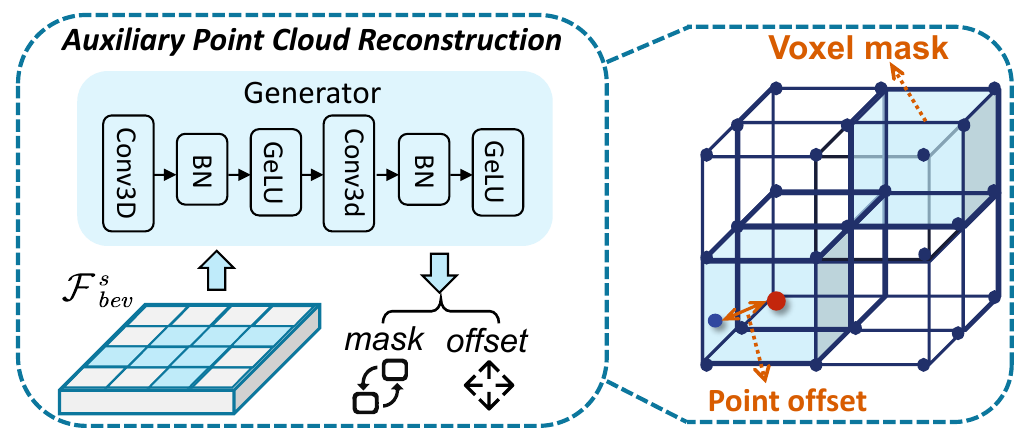}
    \caption{Auxiliary Point Cloud Reconstruction Task. We project $\mathcal{F}_{bev}^{s}$ back into 3D space and decouple it into voxel occupancy prediction and point offset estimation to achieve voxel-level point cloud reconstruction.}
    \label{apcr}
\end{minipage}
\hfill
\begin{minipage}{0.52\columnwidth}
    \centering
    \includegraphics[width=\linewidth]{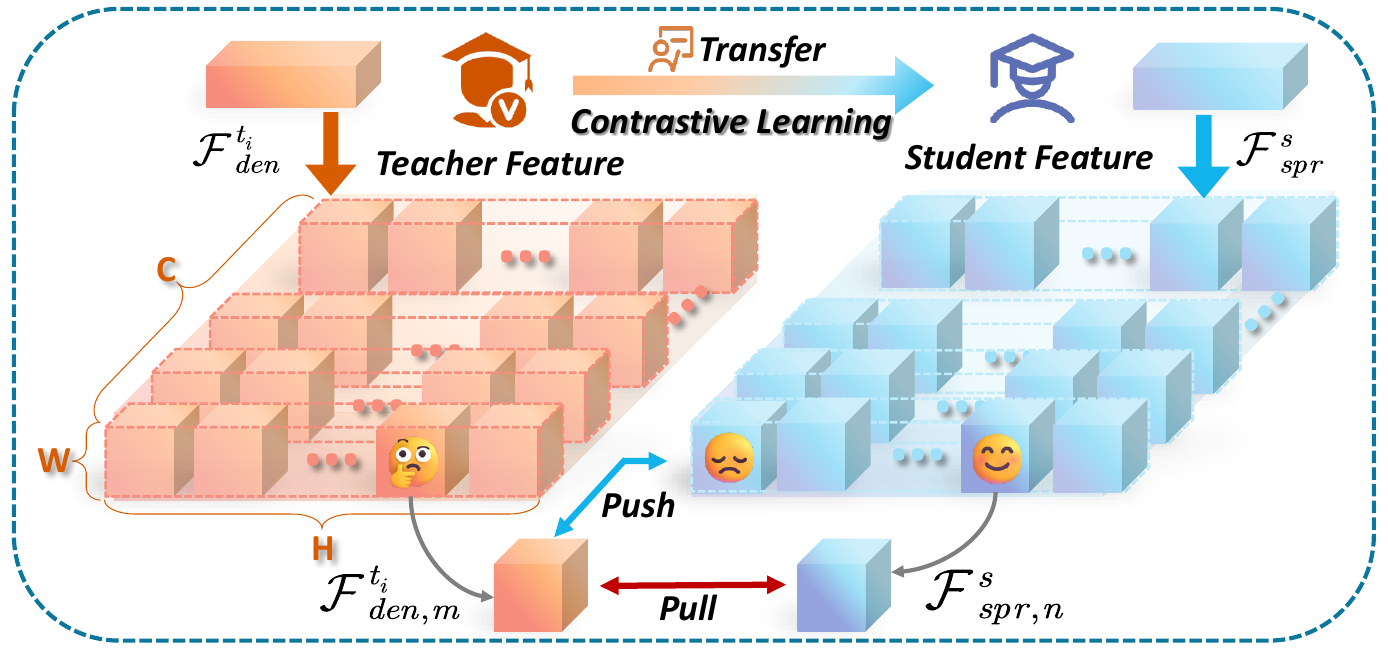}
  \caption{Multi-Teacher Contrastive Learning. We first partition the feature map into non-overlapping patches of equal size, then perform spatial and channel comparisons within each local patch.}
  \label{mtcl}
\end{minipage}
\end{figure}

Since directly recovering dense point clouds incurs substantial computational and memory overhead, inspired by~\cite{Sparse2Dense}, we decompose the auxiliary point cloud reconstruction task into two sub-tasks: voxel occupancy mask $\mathcal{R}_{occ}$ prediction and point offset $\mathcal{R}_{off}$ estimation.  As illustrated in the Fig. \ref{apcr}, APCR first performs dimensionality expansion on the 2D feature $\mathcal{F}_{bev}^{s}$, projecting it back into 3D space, and then processes it through multiple layers of $1 \times 1$ 3D convolution with GELU activation to generate $\mathcal{R}_{occ}$ and $\mathcal{R}_{off}$.

Thus, the reconstructed point cloud can be calculated as:  
\begin{equation}
\begin{aligned}
l^{rec} = (\mathcal{V}_{c} + \mathcal{R}_{off}) \times \mathcal{R}_{occ},
\end{aligned}
\end{equation}
where $ l^{rec}$ is the reconstructed dense point cloud and $\mathcal{V}_{c}$ denotes to the center of voxel.

We supervise the voxel occupancy prediction mask using a binary cross-entropy $H$ loss, and the point offset estimation using an L1 regression loss. They are defined as:
\begin{equation}
\begin{aligned}
\mathcal{L}_{occ}
    &= -\frac{N_{bg}}{N_{fg}} \, H(\mathcal{G}_{occ}, \mathcal{P}_{occ}), 
    \\
\mathcal{L}_{off}
    &= \frac{1}{|N_{fg}|} \sum_{i=1}^{N_{fg}} \| (\mathcal{V}_{c_i} + \mathcal{R}_{off_i}) - O_{gt_i} \| ,
\end{aligned}
\end{equation}
where $\mathcal{G}_{occ}$ denotes the ground-truth occupancy annotation and $\mathcal{P}_{occ}$ denotes the predicted occupancy probability;
$N_{fg}$ and $N_{bg}$ are the numbers of foreground and background voxels; $O_{gt_i}$ represents the ground-truth point cloud of the $i$-th foreground voxel.

The final reconstruction loss is the sum of occupancy loss and offset loss:  
\begin{equation}
\begin{aligned}
\mathcal{L}_{apcr} = \mathcal{L}_{occ} + \mathcal{L}_{off} .
\end{aligned}
\end{equation} 

\subsection{Multi-Teacher Contrastive Distillation}
Multi-agent teachers contain richer contextual and structural knowledge, while the student's single-view limitation creates a \textbf{feature gap} that hinders the direct acquisition of such dense features. To bridge this gap and effectively transfer locally structured and globally salient features from teacher detectors $\mathcal{D}^{t}$ to the target student model $\mathcal{D}^{s}$, we propose the Multi-Teacher Contrastive Distillation (MTCD) strategy.

As shown in Fig.~\ref{mtcl}, we partition each BEV feature map into multiple non-overlapping regions, which are treated as contrastive samples. Unlike label-based contrastive frameworks~\cite{cirkd}, 
the positive and negative pairs are constructed within teacher–student feature correspondences $(\mathcal{F}_{den}^{t}, \mathcal{F}_{spr}^{s})$, based on their spatial alignment. 
Formally, a positive pair is defined for identical spatial indices $(m = n)$ between $(\mathcal{F}_{spr,m}^{s}, \mathcal{F}_{den,n}^{t})$, where $m$ and $n$ denote the spatial locations on the BEV feature map. 
Conversely, negative pairs correspond to index mismatches $(m \neq n)$.

The contrastive loss $\mathcal{L}_{CL}^{t_i}$ for the $i$-th teacher is:
\begin{equation}
\begin{aligned}
\mathcal{L}_{CL}^{t_i} = \frac{1}{N} \sum_{m=1}^{N} -\log \frac{\exp(-d_{m,m}^{t_i} / \tau)}{ \sum_{n=1}^{N} \mathbbm{1}_{m \neq n} \exp(-d_{m,n}^{t_i} / \tau)},
\end{aligned}
\end{equation}
Where $\tau = 0.07$ is the temperature parameter, $N$ denotes the number of contrastive pairs, and $\mathbbm{1}_{m \neq n}$ is an indicator function that equals 1 only when $m \neq n$. 
The feature similarity is measured using the squared Euclidean distance: $d_{m,m}^{t_i} = \|\mathcal{F}_{spr,m}^{s}- \mathcal{F}_{den,m}^{t_i}\|_2^2$ and 
$d_{m,n}^{t_i} = \|\mathcal{F}_{spr,m}^{s}- \mathcal{F}_{den,n}^{t_i}\|_2^2$ . 

To enable the student to capture the teacher’s locally dense and structured knowledge by focusing on contextual cues and position–channel group information, we combine two complementary contrastive modes: Spatial Contrasting and Channel Contrasting.
The detailed formulation is provided in the \textit{\textcolor{magenta}{Appx.}}.

Accordingly, we use $\mathcal{L}_{local}$ to emphasize fine-grained local structural alignment, which is defined as:
\begin{equation}
\begin{aligned}
\mathcal{L}_{local} = \sum_{i=1}^{T} W_i \cdot \mathcal{L}_{CL}^{t_i},
\end{aligned}
\end{equation} 
where $W_i$ is the confidence weight for the $i$-th teacher, based on multi-teacher adaptive distillation mechanism. 

\textbf{Multi-Teacher Adaptive Distillation.} Due to variations in perceptual capabilities across different teacher models under diverse scenarios or samples, direct averaging fusion may introduce noise or even misguide training. To achieve high-quality knowledge transfer, we further designed a Multi-Teacher Adaptive Distillation mechanism. This design dynamically assigns weights based on the teacher model's performance on the current sample. 

Specifically, we use the localization loss of 3D bounding box prediction as the performance indicator. 
For the $i$-th teacher, the localization loss $\mathcal{L}_{loc}^{t_i}$ is computed by Smooth $L_1$ loss between the predicted box $\mathcal{\hat{B}}_{t_i}$ and the ground truth $\mathcal{B}_{gt}$: $\mathcal{L}_{loc}^{t_i} = \text{Smooth}_{L_1} \left( \mathcal{\hat{B}}_{t_i}, \mathcal{B}_{gt} \right).$

Then, the confidence weight $W_i$ is obtained by the localization loss across all teachers:
\begin{equation}
W_i = \frac{\exp(-\alpha \cdot \mathcal{L}_{loc}^{t_i})}{\sum_{j=1}^{T} \exp(-\alpha \cdot \mathcal{L}_{loc}^{t_j})},
\label{eq:weight}
\end{equation}
where $\alpha=0.8$ is a scaling factor that controls the sensitivity of the confidence to the loss. This weighting scheme prioritizes teachers with higher confidence.

Moreover, to further guide the student $\mathcal{D}^s$ to approximate the comprehensive dense BEV feature $\hat{\mathcal{F}}_{den,k}^{t}$ of teachers $\mathcal{D}^t$ in global level, we employ a multi-teacher adaptive distillation approach. Specially, we consider $T$ pre-trained teacher models $\mathcal{D}^{t_i}$, which generate dense features. For the $k$-th layer, let $\mathcal{F}_{den,k}^{t_i} = \mathcal{D}^{t_i}(l^{coop})$ denote the dense BEV features of the $i$-th teacher. We fused these features into $\hat{\mathcal{F}}_{den,k}^{t}$ by:
\begin{equation}
\hat{\mathcal{F}}_{den,k}^{t} = \sum_{i=1}^{T} W_i \cdot \mathcal{F}_{den,k}^{t_i},
\label{eq:fusion}
\end{equation}
where $W_i$ is the confidence weight for the $i$-th teacher, which is computed based on the detection performance of the teacher. 

Finally, we employ KL-divergence feature distillation as:
\begin{equation}
\mathcal{L}_{global} = \sum_{k=1}^K \beta^k_{kd} \cdot \mathcal{D}_{KL} \left( \rho(\mathcal{F}^{s}_{spr,k}) \parallel \rho(\hat{\mathcal{F}}_{den,k}^{t}) \right),
\label{eq:feature_distill}
\end{equation}
where $\rho$ denotes softmax function, and $\mathcal{D}_{\mathrm{KL}}(\pi \| p) = \mathbb{E}_{\pi(x)} [ \log \frac{\pi(x)}{p(x)} ]$
is the Kullback-Leibler divergence, and the hyperparameter $\beta_{kd}^k \in \{0.2,0.3,0.5 \}$ to balance weights across different feature layers, which set same to~\cite{MKD-cooper}.

\subsection{Loss Function} 

We train our M2S using the following loss function: 
\begin{equation}
\begin{aligned}
\mathcal{L}_{M2S} = \gamma_{cls} \mathcal{L}_{cls} + \gamma_{loc} \mathcal{L}_{loc} + \gamma_{apcr} \mathcal{L}_{apcr}\\
 + \gamma_{global} \mathcal{L}_{global}  + \gamma_{local} \mathcal{L}_{local},
\end{aligned}
\end{equation} 
where the hyperparameters $\gamma_{cls}$, $\gamma_{loc}$, $\gamma_{apcr}$, $\gamma_{global}$ and $\gamma_{local}$  balance the contributions of different losses, and $\mathcal{L}_{cls}$ denotes the classification loss implemented with focal loss, $\mathcal{L}_{loc}$ represents the bbox regression loss using Smooth $L_1$.

\section{Experiments}

\subsection{Datasets and Evaluation Metrics}
\textbf{Datasets.} 
We conduct comprehensive experiments on two large-scale real-world datasets, V2V4Real~\cite{v2v4real} and DAIR-V2X~\cite{dair-v2x}, as well as a simulated dataset, V2XSet~\cite{V2X-ViT}. 
We provide more datasets information in the \textit{\textcolor{magenta}{Appx.}}.

\textbf{Evaluation Metrics.} 
We adopt the {average precision (AP)} as the main indicators for evaluating 3D object detection performance. Additionally, we record the {delay} and {frames per second (FPS)} to measure the communication latency and inference efficiency, respectively. And we report the number of {floating-point operations (FLOPs)} to assess the computational complexity during inference. Furthermore, we use single NVIDIA RTX 3090 GPU to report training time (hours) and GPU memory usage (MB) and discuss the training cost.

\begin{table*}[!t]
\caption{Results on the V2XSet. We report the performance of different models on the V2XSet with \textbf{single-agent input}. Improvements are highlighted in \textbf{bold}.}
\label{tab:V2XSet}
\centering
\resizebox{\textwidth}{!}{
\setlength{\arrayrulewidth}{0.6pt}
\begin{tabular}{c|c|ccc|ccc }
\hline
\multirow{2}{*}{\textbf{Models}} & \multirow{2}{*}{\textbf{Publication}} & \multicolumn{3}{c|}{\textbf{3D mAP$@$Validation}} & \multicolumn{3}{c}{\textbf{3D mAP$@$Testing}} \\
 &  & \textbf{IoU=0.3} & \textbf{IoU=0.5} & \textbf{IoU=0.7} & \textbf{IoU=0.3} & \textbf{IoU=0.5} & \textbf{IoU=0.7} \\
\hline
1.AttFuse & \multirow{3}{*}{ICRA 2022}  & 78.89 & 76.72 & 68.10 & 75.67 & 73.72 & 58.06 \\
2.AttFuse w/ M2S &   & 85.20 & 82.83 & 73.73 & 82.15 & 79.26 & 66.70 \\
{Improvement (2-1)} &   & \cellcolor{gray!10}\textbf{{+6.31}} & \cellcolor{gray!10}\textbf{{+6.11}} & \cellcolor{gray!10}\textbf{{+5.63}} & \cellcolor{gray!10}\textbf{{+6.48}} & \cellcolor{gray!10}\textbf{{+5.54}} & \cellcolor{gray!10}\textbf{{+8.64}} \\
\hline
1.V2X-ViT & \multirow{3}{*}{ECCV 2022}  & 78.20 & 76.84 & 67.51 & 78.86 & 76.17 & 61.90 \\
2.V2X-ViT w/ M2S &   & 83.97 & 82.45 & 72.49 & 83.91 & 81.02 & 65.25 \\
{Improvement (2-1)} &   & \cellcolor{gray!10}\textbf{{+5.77}} & \cellcolor{gray!10}\textbf{{+5.61}} & \cellcolor{gray!10}\textbf{{+4.98}} & \cellcolor{gray!10}\textbf{{+5.05}} & \cellcolor{gray!10}\textbf{{+4.85}} & \cellcolor{gray!10}\textbf{{+3.35}} \\
\hline
1.Where2comm & \multirow{3}{*}{NeurIPS 2022}  & 80.22 & 78.23 & 68.44 & 79.21 & 76.24 & 62.18 \\
2.Where2comm w/ M2S &   & 84.99 & 83.26 & 75.15 & 84.42 & 82.32 & 70.28 \\
{Improvement (2-1)} &   & \cellcolor{gray!10}\textbf{{+4.77}} & \cellcolor{gray!10}\textbf{{+5.03}} & \cellcolor{gray!10}\textbf{{+6.71}} & \cellcolor{gray!10}\textbf{{+5.21}} & \cellcolor{gray!10}\textbf{{+6.08}} & \cellcolor{gray!10}\textbf{{+8.10}} \\
\hline
1.CoAlign & \multirow{3}{*}{ICRA 2023}  & 80.60 & 78.63 & 69.74 & 80.62 & 78.02 & 65.07 \\
2.CoAlign w/ M2S &   & 88.11 & 86.27 & 77.15 & 86.49 & 83.88 & 71.66 \\
{Improvement (2-1)} &   & \cellcolor{gray!10}\textbf{{+7.51}} & \cellcolor{gray!10}\textbf{{+7.64}} & \cellcolor{gray!10}\textbf{{+7.41}} & \cellcolor{gray!10}\textbf{{+5.87}} & \cellcolor{gray!10}\textbf{{+5.86}} & \cellcolor{gray!10}\textbf{{+6.59}} \\
\hline
1.CoSDH & \multirow{3}{*}{CVPR 2025}  & 80.74 & 78.53 & 67.91 & 79.52 & 76.75 & 62.05 \\
2.CoSDH w/ M2S &   & 87.00 & 84.18 & 75.13 & 84.81 & 81.39 & 66.50 \\
{Improvement (2-1)} &   & \cellcolor{gray!10}\textbf{{+6.26}} & \cellcolor{gray!10}\textbf{{+5.65}} & \cellcolor{gray!10}\textbf{{+7.22}} & \cellcolor{gray!10}\textbf{{+5.29}} & \cellcolor{gray!10}\textbf{{+4.64}} & \cellcolor{gray!10}\textbf{{+4.45}} \\
\hline
\hline
1.PillarNeSt & \multirow{3}{*}{TIV 2024}  & 79.14 & 77.66 & 67.85 & 77.05 & 74.29 & 59.83 \\
2.PillarNeSt w/ M2S &   & 85.94 & 84.09 & 71.82 & 84.14 & 81.53 & 64.16 \\
{Improvement (2-1)} &   & \cellcolor{gray!10}\textbf{{+6.80}} & \cellcolor{gray!10}\textbf{{+6.43}} & \cellcolor{gray!10}\textbf{{+3.97}} & \cellcolor{gray!10}\textbf{{+7.09}} & \cellcolor{gray!10}\textbf{{+7.24}} & \cellcolor{gray!10}\textbf{{+4.33}} \\
\hline
\end{tabular}%
}
\end{table*}

\subsection{Baselines Selection and Experimental Details} 
\textbf{Baselines Selection.}
We use traditional ego-only methods, for instance, PillarNeSt~\cite{PillarNeSt} to validate our M2S framework and also run the collaborative baselines, including AttFuse~\cite{attfuse}, V2X-ViT~\cite{V2X-ViT}, Where2comm~\cite{where2comm}, CoAlign~\cite{coalign}, and CoSDH~\cite{cosdh}, with ego-only input to examine performance in realistic ego-only perception scenarios. Overall, our experimental setup was designed to fully validate the performance of our M2S across different perception methods.

\textbf{Experimental Details.} 
We use the ego agent for evaluation, following the settings of~\cite{attfuse}. We employ three pre-trained models as teachers: AttFuse~\cite{attfuse}, V2X-ViT~\cite{V2X-ViT}, and CoSDH~\cite{cosdh}. We implement a teacher network that receives only the multi-agent input and a student network that receives only the ego-agent input. During training, the student network is constructed following the full collaborative architecture, including the fusion modules, but receives only the ego-agent input. During inference, only the student detector, which takes only the ego-agent input, is used for prediction to ensure a fair comparison. We provide more experimental details in the \textit{\textcolor{magenta}{Appx.}}.

\subsection{Performance Evaluation}
\textbf{Results on the V2XSet dataset.} 
Table~\ref{tab:V2XSet} compares several baselines with our M2S framework. Both the collaborative perception methods and ego-only
perception method~\cite{PillarNeSt} baselines exhibit clear performance gains with our M2S. Notably, on the test set, M2S improves 3D mAP (IoU$=$0.7) by 8.64\%, 3.35\%, 8.10\%, 6.59\%, 4.45\%, and 4.33\% for AttFuse~\cite{attfuse}, V2X-ViT~\cite{V2X-ViT}, Where2comm~\cite{where2comm}, CoAlign~\cite{coalign}, CoSDH~\cite{cosdh}, and PillarNeSt~\cite{PillarNeSt}, respectively. Most importantly, these gains are achieved with no extra communication overhead, validating our M2S's effectiveness and practicality.

\begin{table*}[!t]
\caption{Results on the V2V4Real and DAIR-V2X. We show the performance of different models with \textbf{single-agent input}. Improvements are highlighted in \textbf{bold}.}
\label{tab:V2V4Real}
\centering
\resizebox{\textwidth}{!}{
\setlength{\arrayrulewidth}{0.6pt}
\renewcommand{\arraystretch}{1.2}
\begin{tabular}{c|ccc|ccc|cccc}
\hline
\multirow{2}{*}{\textbf{Models}} & \multicolumn{3}{c|}{\textbf{V2V4Real Val}} & \multicolumn{3}{c|}{\textbf{V2V4Real Test}} & \multicolumn{3}{c}{\textbf{DAIR-V2X Val}} \\
 & \textbf{AP$@$0.3} & \textbf{AP$@$0.5} & \textbf{AP$@$0.7} & \textbf{AP$@$0.3} & \textbf{AP$@$0.5} & \textbf{AP$@$0.7} & \textbf{AP$@$0.3} & \textbf{AP$@$0.5} & \textbf{AP$@$0.7} \\
\hline
1.AttFuse & 49.22 & 44.80 & 25.69 & 44.64 & 40.66 & 21.49 & 58.97 & 49.70 & 33.25 \\
2.AttFuse~w/ M2S & 57.16 & 47.44 & 26.57 & 53.85 & 45.76 & 26.88 & 61.02 & 52.16 & 39.86 \\
{Improvement (2-1)} & \cellcolor{gray!10}\textbf{+7.94} & \cellcolor{gray!10}\textbf{+2.64} & \cellcolor{gray!10}\textbf{+0.88} & \cellcolor{gray!10}\textbf{+9.21} & \cellcolor{gray!10}\textbf{+5.10} & \cellcolor{gray!10}\textbf{+5.39} & \cellcolor{gray!10}\textbf{+2.05} & \cellcolor{gray!10}\textbf{+2.46} & \cellcolor{gray!10}\textbf{+6.61} \\
\hline
1.V2X-ViT & 52.00 & 46.98 & 25.30 & 46.59 & 41.40 & 19.56 & 60.71 & 51.06 & 35.08 \\
2.V2X-ViT~w/ M2S & 58.78 & 47.44 & 26.77 & 54.66 & 44.49 & 20.85 & 60.65 & 52.77 & 39.77 \\
{Improvement (2-1)} & \cellcolor{gray!10}\textbf{+6.78} & \cellcolor{gray!10}\textbf{+0.46} & \cellcolor{gray!10}\textbf{+1.47} & \cellcolor{gray!10}\textbf{+8.07} & \cellcolor{gray!10}\textbf{+3.09} & \cellcolor{gray!10}\textbf{+1.29} & \cellcolor{gray!10}{-0.06} & \cellcolor{gray!10}\textbf{+1.71} & \cellcolor{gray!10}\textbf{+4.69} \\
\hline
1.Where2comm & 46.80 & 43.57 & 27.03 & 47.33 & 42.08 & 25.17 & 59.42 & 51.10 & 38.54 \\
2.Where2comm~w/ M2S & 57.04 & 44.72 & 25.55 & 55.50 & 47.07 & 26.79 & 60.84 & 52.24 & 39.84 \\
{Improvement (2-1)} & \cellcolor{gray!10}\textbf{+10.24} & \cellcolor{gray!10}\textbf{+1.15} & \cellcolor{gray!10}{-1.48} & \cellcolor{gray!10}\textbf{+8.17} & \cellcolor{gray!10}\textbf{+4.99} & \cellcolor{gray!10}\textbf{+1.62} & \cellcolor{gray!10}\textbf{+1.42} & \cellcolor{gray!10}\textbf{+1.14} & \cellcolor{gray!10}\textbf{+1.30} \\
\hline
1.CoAlign & 46.67 & 41.75 & 23.77 & 48.54 & 40.94 & 21.48 & 61.26 & 51.26 & 37.86 \\
2.CoAlign~w/ M2S & 55.16 & 45.45 & 24.13 & 54.27 & 46.84 & 23.74 & 62.77 & 53.14 & 38.75 \\
{Improvement (2-1)} & \cellcolor{gray!10}\textbf{+8.49} & \cellcolor{gray!10}\textbf{+3.70} & \cellcolor{gray!10}\textbf{+0.36} & \cellcolor{gray!10}\textbf{+5.73} & \cellcolor{gray!10}\textbf{+5.90} & \cellcolor{gray!10}\textbf{+2.26} & \cellcolor{gray!10}\textbf{+1.51} & \cellcolor{gray!10}\textbf{+1.94} & \cellcolor{gray!10}\textbf{+0.95} \\
\hline
1.CoSDH & 46.04 & 42.85 & 25.07 & 46.79 & 43.20 & 23.11 & 57.50 & 48.57 & 34.37 \\
2.CoSDH~w/ M2S & 55.98 & 46.25 & 24.42 & 56.54 & 47.91 & 25.53 & 60.19 & 51.67 & 37.33 \\
{Improvement (2-1)} & \cellcolor{gray!10}\textbf{+9.94} & \cellcolor{gray!10}\textbf{+3.40} & \cellcolor{gray!10}{-0.65} & \cellcolor{gray!10}\textbf{+9.75} & \cellcolor{gray!10}\textbf{+4.71} & \cellcolor{gray!10}\textbf{+2.42} & \cellcolor{gray!10}\textbf{+2.69} & \cellcolor{gray!10}\textbf{+3.10} & \cellcolor{gray!10}\textbf{+2.96} \\
\hline
\hline
1.PillarNeSt & 50.61 & 45.36 & 21.70 & 47.90 & 40.20 & 16.18 & 57.30 & 48.90 & 33.46 \\
2.PillarNeSt~w/ M2S & 52.19 & 45.28 & 26.03 & 53.99 & 43.55 & 25.24 & 60.36 & 51.85 & 38.17 \\
{Improvement (2-1)} & \cellcolor{gray!10}\textbf{+1.58} & \cellcolor{gray!10}{-0.08} & \cellcolor{gray!10}\textbf{+4.33} & \cellcolor{gray!10}\textbf{+6.09} & \cellcolor{gray!10}\textbf{+3.35} & \cellcolor{gray!10}\textbf{+9.06} & \cellcolor{gray!10}\textbf{+3.06} & \cellcolor{gray!10}\textbf{+2.95} & \cellcolor{gray!10}\textbf{+4.71} \\
\hline
\end{tabular}%
}
\end{table*}

\textbf{Results on the real-world datasets.} 
To further verify the effectiveness and generalization of our M2S in real-world scenarios, we conduct experiments on V2V4Real~\cite{v2v4real} and DAIR-V2X~\cite{dair-v2x}. As with  V2XSet~\cite{V2X-ViT}, both collaborative~\cite{attfuse,V2X-ViT,where2comm,coalign,cosdh} and ego-only~\cite{PillarNeSt} baselines gain consistently. Specifically, On V2V4Real test set at AP$@$0.7, all baselines consistently improve with M2S, with PillarNeSt achieving the largest gain of $9.06$, followed by AttFuse ($+5.39$), CoSDH ($+2.42$), and CoAlign ($+2.26$). 
A similar trend is observed on the DAIR-V2X, where M2S consistently enhances performance across multiple models, confirming its robustness across diverse environments. 

\begin{table}[!t]
\caption{Ablation study of M2S components. Results report 3D mAP (IoU $=$ 0.3/0.5/0.7) on the V2XSet. Here, ST denotes single-teacher distillation (AttFuse only), MTCD$^\dagger$ indicates multi-teacher adaptive distillation, which refers to global-level feature distillation. MLFE is multi-level feature enhancement, APCR refers to auxiliary point cloud reconstruction, and MTCD$^\ddagger$ is multi-teacher contrastive learning, which refers to local-level feature distillation. Best  are highlighted in \textbf{bold}.}
\label{tab:module_ablation}
\centering
\resizebox{1.0\textwidth}{!}{%
\begin{tabular}{@{}c|ccccc|ccc|ccc@{}}
\toprule
\multirow{2}{*}{\textbf{Row}} &\multicolumn{5}{c|}{\textbf{Module}} & \multicolumn{3}{c|}{\textbf{Validation}} & \multicolumn{3}{c}{\textbf{Testing}} \\
& \textbf{ST} & \textbf{MTCD$^\dagger$} & \textbf{MLFE} & \textbf{APCR} & \textbf{MTCD$^\ddagger$} & \textbf{AP$@$0.3} & \textbf{AP$@$0.5} & \textbf{AP$@$0.7} & \textbf{AP$@$0.3} & \textbf{AP$@$0.5} & \textbf{AP$@$0.7} \\
\midrule
1 &- & - & - & - & - & 78.89 & 76.72 & 68.10 & 75.67 & 73.72 & 58.06 \\
2 & $\checkmark$ & - & - & - & - & 82.83 & 80.17 & 68.54 & 78.10 & 74.77 & 58.29 \\
3 & - & $\checkmark$ & - & - & - & 84.10 & 81.91 & 71.64 & 81.98 & 78.55 & 62.76 \\
4 & - & $\checkmark$ & $\checkmark$ & - & - & 84.31 & 82.27 & 72.85 & 82.15 & \textbf{79.44} & 64.58 \\
5 &- & $\checkmark$ & $\checkmark$ & $\checkmark$ & - & 84.91 & 82.75 & 73.08 & 81.80 & 79.32 & 66.50 \\
6 & - & $\checkmark$ & $\checkmark$ & $\checkmark$ & $\checkmark$ & \textbf{85.20} & \textbf{82.83} & \textbf{73.73} & \textbf{82.15} & 79.26 & \textbf{66.70} \\
\bottomrule
\end{tabular}%
}
\end{table}

\subsection{Ablation Study}
Table~\ref{tab:module_ablation} ablates each module in M2S. We adopt AttFuse~\cite{attfuse}, V2X-ViT~\cite{V2X-ViT}, and CoSDH~\cite{cosdh} as teacher models, with AttFuse that receives single-agent input as the student. Introducing single-teacher distillation (ST) alone yields a slight improvement in AP$@$0.7 from 58.06\% to 58.29\%. Replacing it with the multi-teacher adaptive distillation module  (row 3) significantly boosts performance to 62.76\%, demonstrating the benefit of leveraging multiple teachers. Adding the multi-level feature enhancement (MLFE) and auxiliary point cloud reconstruction (APCR) modules further increases AP$@$0.7 to 66.50\%. Finally, incorporating multi-teacher contrastive learning (row 6) achieves the best result of 66.70\%, confirming the complementary effectiveness of all components in M2S.

\subsection{Analysis Experiments}

\begin{figure}[!t]
  \centering
  \begin{minipage}{0.48\textwidth}
    \centering
    \includegraphics[width=\linewidth]{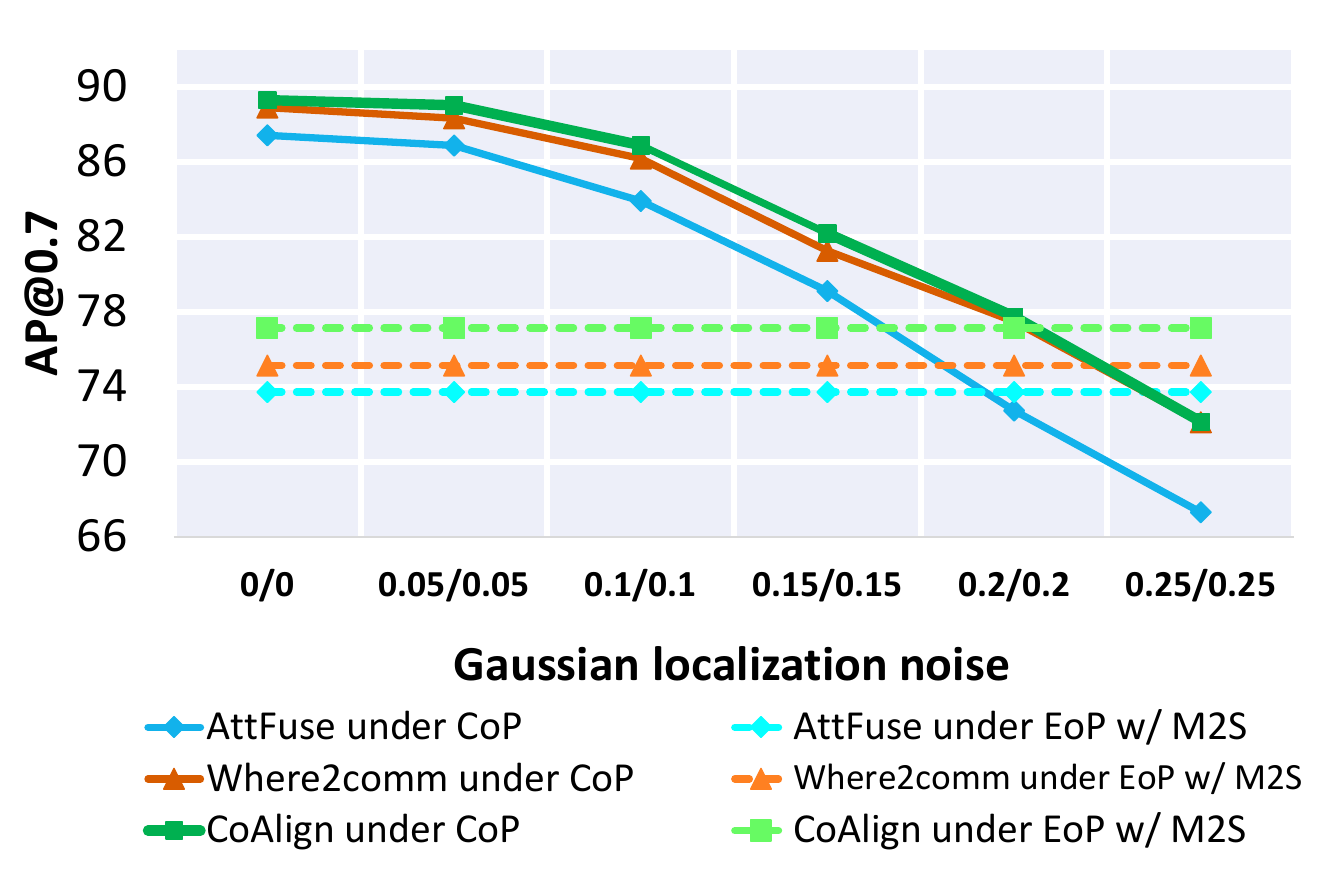}
    \captionof{figure}{Comparison performance of noise robustness under Co-Perception (CoP) and Eo-perception (EoP) w/ M2S modes. We report the 3D mAP (IoU=0.7) of AttFuse, Where2comm and CoAlign under diffent gaussian noise on the V2XSet.}
    \label{fig:noise}
  \end{minipage}
  \hfill
  \begin{minipage}{0.5\textwidth}
    \centering
    \captionof{table}{Comparison of computational complexity, inference speed, and communication latency under CoP and EoP w/ M2S. FLOPs (G) denotes model complexity, FPS (fps) is the inference speed, and Delay (ms) is communication latency.}
    \label{tab:comm}
    \resizebox{\linewidth}{!}{%
      \begin{tabular}{c|c|c|c|c}
        \hline
        \textbf{Methods} & \textbf{Type} & \textbf{FLOPs $\downarrow$} & \textbf{FPS $\uparrow$} & \textbf{Delay $\downarrow$} \\
        \hline
        \multirow{2}{*}{AttFuse} & CoP & 143.8 & 10.3 & 266.6 \\
         & EoP w/M2S & 92.8 & 20.7 & 0 \\
        \hline
        \multirow{2}{*}{V2X-ViT} & CoP & 315.9 & 6.6 & 261.6 \\
         & EoP w/M2S & 205.3 & 9.2 & 0 \\
        \hline
        \multirow{2}{*}{Where2comm} & CoP & 415.9 & 12.1 & 281.9 \\
         & EoP w/M2S & 185.5 & 16.0 & 0 \\
        \hline
        \multirow{2}{*}{CoAlign} & CoP & 138.1 & 15.7 & 276.7 \\
         & EoP w/M2S & 92.9 & 17.7 & 0 \\
        \hline
        \multirow{2}{*}{CoSDH} & CoP & 428.6 & 12.9 & 260.5 \\
         & EoP w/M2S & 189.7 & 14.3 & 0 \\
        \hline
        \multirow{2}{*}{PillarNeSt} & CoP & 893.8 & 10.6 & 274.5 \\
         & EoP w/M2S & 391.8 & 13.9 & 0 \\
        \hline
      \end{tabular}%
    }
  \end{minipage}
\end{figure}

\textbf{Comparison of noise robustness.}
Following the noise injection settings from ~\cite{where2comm} and ~\cite{V2X-ViT}, we applied Gaussian noise  $\zeta$ from ($ 0.0\,\text{m}, 0.0^\circ$) to ($ 0.25\,\text{m}, 0.25^\circ$) to simulate localization errors under complex scenarios. 
As shown in Fig.~\ref{fig:noise}, collaborative perception methods suffer severe performance degradation as errors accumulate. In contrast, when the error is within ($ 0.25\,\text{m}, 0.25^\circ$), our M2S maintains consistent performance by relying solely on ego-agent inference, showing that M2S can inherently mitigate the negative impact of localization errors.

\textbf{Comparison of computational complexity, inference speed, and communication latency.}
Table~\ref{tab:comm} compares Co-Perception (CoP) and Ego-Perception (EoP) with our M2S \textbf{during inference}. The CoP receives multi-agent input, while the EoP only receives the ego. Overall, M2S notably reduces computational complexity, accelerates inference, and eliminates communication latency. Specifically, M2S reduces FLOPs by 35.43\% and doubles FPS for AttFuse~\cite{attfuse}. 
Since our M2S does not require external communication, which eliminates transmission delays, whereas collaboration methods suffer from 260 to 282 ms latency. In short, these results confirm M2S's practicality and deployment potential.

\begin{table}[!t]
\caption{Comparison of different teacher combinations. We compared five teacher combinations and uniformly used ~\cite{attfuse} as student. Best in \textbf{bold} and second in \underline{underline}.}
\label{tab:teacher_model_performance}
\centering
\resizebox{0.8\textwidth}{!}{%
\begin{tabular}{c|ccc|c|c}
\hline
\multirow{2}{*}{\textbf{Teacher Combinations}} & \multicolumn{3}{c|}{\textbf{Performance}} & \textbf{Train} & \textbf{GPU} \\
 & \textbf{AP$@$0.3} & \textbf{AP$@$0.5} & \textbf{AP$@$0.7} & \textbf{Time (h)} & \textbf{Mem.(M)}  \\
\hline
AttFuse (Single teacher) & \underline{84.34} & 82.35 & 73.26 & 7.54 & 7640 \\
\hline
AttFuse+CoSDH           & 83.95 & 82.20 & \underline{73.69} & 8.05  & 7724 \\
AttFuse+V2X-ViT         & 83.74 & 81.45 & 73.41 & 12.13 & 10730 \\
CoSDH+V2X-ViT      & 84.25 & \underline{82.74} & 73.30 & 12.10  & 10348 \\
\hline
AttFuse+V2X-ViT+CoSDH   & \textbf{85.20} & \textbf{82.83} & \textbf{73.73} & 12.69 & 10992 \\
\hline
\end{tabular}%
}
\end{table}

\begin{table*}[!t]
\centering
\caption{Performance comparison of different methods under various range settings. We report the reults of V2Xset with \textbf{single-agent input}. Best in \textbf{bold}.}
\label{tab:range_performance}
\resizebox{\textwidth}{!}{%
\setlength{\arrayrulewidth}{0.6pt}
\small
\begin{tabular}{c|cc|cc|cc|cc}
\hline
\multirow{2}{*}{\textbf{Methods}} & \multicolumn{2}{c|}{\textbf{All Range}} & \multicolumn{2}{c|}{\textbf{Range [0,30) ($m$)}} & \multicolumn{2}{c|}{\textbf{Range[30,50) ($m$)}} & \multicolumn{2}{c}{\textbf{Range[50,+inf) ($m$)}} \\
\cline{2-9}
 & \textbf{AP$@$0.5} & \textbf{AP$@$0.7} & \textbf{AP$@$0.5} & \textbf{AP$@$0.7}& \textbf{AP$@$0.5} & \textbf{AP$@$0.7} & \textbf{AP$@$0.5} & \textbf{AP$@$0.7} \\
\hline
AttFuse & 76.72 & 68.10 & 84.34 & 81.08 & 76.31 & 67.41 & 58.43 & 39.63 \\
AttFuse w/M2S & \textbf{82.83} & \textbf{73.73} & \textbf{95.29} & \textbf{92.49} & \textbf{78.13} & \textbf{68.76} & \textbf{60.46} & \textbf{39.97} \\
\hline
Where2comm & 78.23 & 68.44 & 84.55 & 81.70 & \textbf{78.65} & 68.04 & 61.93 & 39.32 \\
Where2comm w/M2S & \textbf{83.26} & \textbf{75.15} & \textbf{95.59} & \textbf{93.23} & 78.10 & \textbf{69.66} & \textbf{62.36} & \textbf{42.60} \\
\hline
CoAlign & 78.63 & 69.74 & 85.08 & 82.15 & 78.31 & 68.95 & 62.93 & 42.31 \\
CoAlign w/M2S & \textbf{86.27} & \textbf{77.15} & \textbf{96.10} & \textbf{93.76} & \textbf{80.88} & \textbf{72.20} & \textbf{71.06} & \textbf{48.54} \\
\hline
\end{tabular}
}
\end{table*}

\textbf{Comparison of teacher combinations and discussion of trade-offs.}
As shown in Table~\ref{tab:teacher_model_performance}, the combination AttFuse+V2X-ViT+CoSDH achieves the highest performance, which comes with substantial training overhead, requiring more training time and higher GPU memory compared to the single-teacher. Notably, other high-cost combinations, like AttFuse+V2X-ViT, fail to deliver comparable accuracy, while the CoSDH+V2X-ViT offers competitive but still inferior results with similar resource consumption. Given its superior accuracy and acceptable training cost, we select AttFuse+V2X-ViT+CoSDH as the final teacher configuration for subsequent experiments.

\textbf{Comparison of performance across distance ranges.}
Table~\ref{tab:range_performance} shows that integrating M2S consistently improves all methods across all distance ranges. CoAlign achieves the largest overall gain, with AP$@$0.7 rising from 69.74 to 77.15. In the short range [0,30), M2S boosts AP$@$0.7 by over 10 points for AttFuse, CoAlign, and Where2comm, indicating enhanced local feature modeling. Gains persist in the mid-range [30,50) and extend to the challenging far-range [50,+inf), where CoAlign’s AP$@$0.5 increases from 62.93 to 71.06. These results demonstrate M2S’s effectiveness across all distance conditions.

\subsection{Qualitative Analysis}
\begin{figure}[!t]
\centering
\begin{minipage}{0.49\columnwidth}
    \centering
    \includegraphics[width=\linewidth]{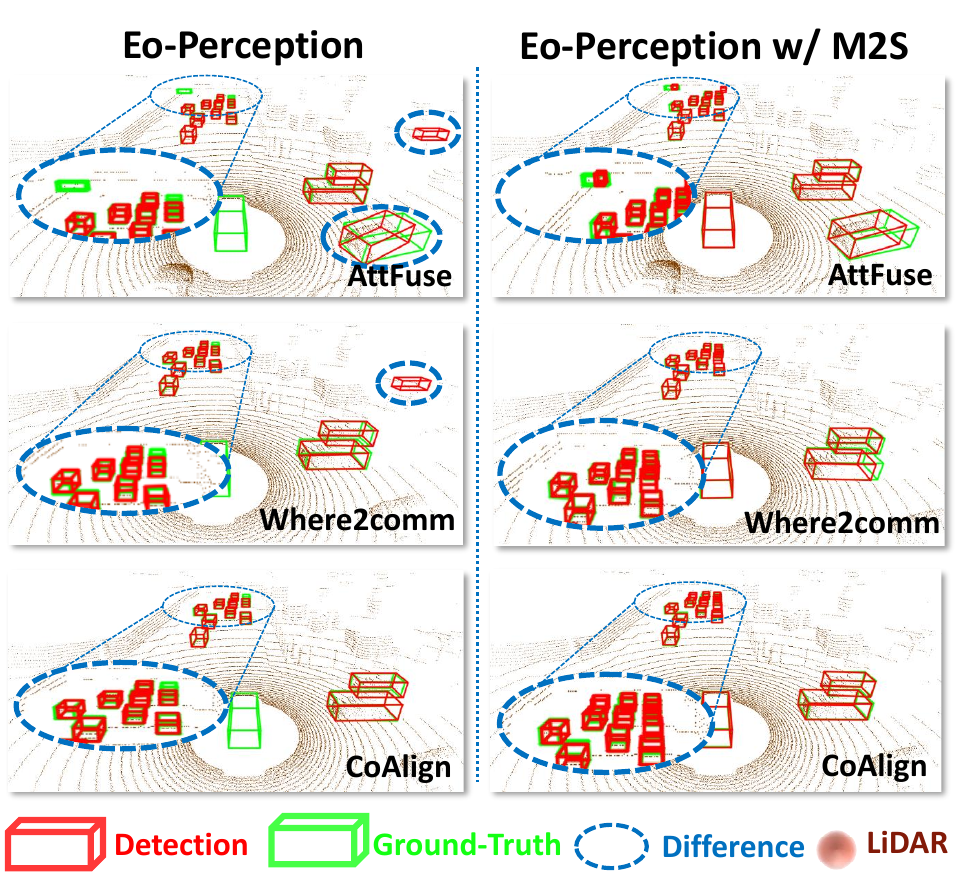}
    \caption{ Visualization of detection results in the Field of view (FOV) on the V2XSet. 
    }
    \label{fig:fov}
\end{minipage}
\hfill
\begin{minipage}{0.5\columnwidth}
    \centering
    \includegraphics[width=\linewidth]{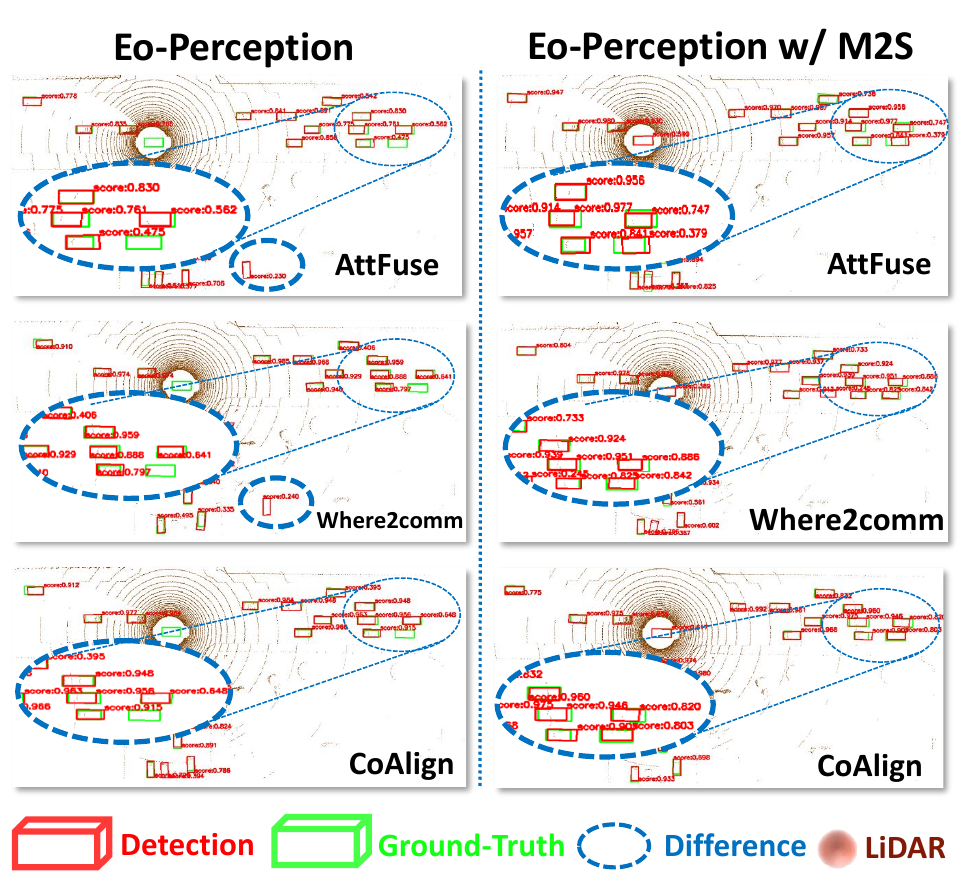}
    \caption{Visualization of detection results in the Bird's Eye View (BEV) on the V2XSet. 
    }
    \label{fig:bev}
\end{minipage}
\end{figure}

To demonstrate M2S's efficacy in transferring dense multi-teacher knowledge to sparse students, we show FOV and BEV results on V2XSet with and without M2S in Figs.~\ref{fig:fov} and~\ref{fig:bev}. M2S achieves more comprehensive and precise detection. Without M2S (Eo-Perception alone), students frequently miss or misclassify objects, especially distant or in dense areas. M2S significantly mitigates these issues, effectively distilling collaborative knowledge and enhancing student performance.

\section{Conclusion}
In this paper, we analyze the inherent advantages and limitations of Co-Perception and Eo-Perception, and explore the new \textbf{C2E} (\textbf{C}o-Perception \textbf{to} \textbf{E}o-Perception) paradigm. Through the \textbf{M}ulti-\textbf{to}-\textbf{S}ingle (\textbf{M2S}) agent contrastive knowledge distillation framework, we achieve state-of-the-art performance on 3D object detection without increasing any communication and inference overhead. Extensive experiments on V2XSet, V2V4Real, and DAIR-V2X datasets demonstrate the effectiveness and generalizability of our M2S framework. In conclusion, the C2E paradigm and M2S framework can provide a new technical path for the reliable deployment of autonomous driving perception systems in real-world scenarios.

\textbf{Limitation and Future Work.} Our current M2S framework is designed for LiDAR-based detection, and extending it to multi-modal sensing (e.g., LiDAR-camera fusion) remains a promising direction. Additionally, the adaptive teacher weighting currently relies on ground-truth labels during training; future work could explore a GT-free multi-teacher voting mechanism to identify and suppress outlier teachers without annotation-based supervision.

\section*{Acknowledgements}
This work was supported in part by the National Natural Science Foundation of China (No.42571514).


%
%
\bibliographystyle{splncs04}
\bibliography{main}

@inproceedings{L4dr,
  title={L4dr: Lidar-4dradar fusion for weather-robust 3d object detection},
  author={Huang, Xun and Xu, Ziyu and Wu, Hai and Wang, Jinlong and Xia, Qiming and Xia, Yan and Li, Jonathan and Gao, Kyle and Wen, Chenglu and Wang, Cheng},
  booktitle={Proceedings of the AAAI conference on artificial intelligence},
  volume={39},
  pages={3806--3814},
  year={2025}
}

@article{yang2023how2comm,
  title={How2comm: Communication-efficient and collaboration-pragmatic multi-agent perception},
  author={Yang, Dingkang and Yang, Kun and Wang, Yuzheng and Liu, Jing and Xu, Zhi and Yin, Rongbin and Zhai, Peng and Zhang, Lihua},
  journal={Advances in Neural Information Processing Systems},
  volume={36},
  pages={25151--25164},
  year={2023}
}

@inproceedings{yang2023what2comm,
  title={What2comm: Towards communication-efficient collaborative perception via feature decoupling},
  author={Yang, Kun and Yang, Dingkang and Zhang, Jingyu and Wang, Hanqi and Sun, Peng and Song, Liang},
  booktitle={Proceedings of the 31st ACM international conference on multimedia},
  pages={7686--7695},
  year={2023}
}

@inproceedings{hinted,
  title={HINTED: Hard Instance Enhanced Detector with Mixed-Density Feature Fusion for Sparsely-Supervised 3D Object Detection},
  author={Xia, Qiming and Ye, Wei and Wu, Hai and Zhao, Shijia and Xing, Leyuan and Huang, Xun and Deng, Jinhao and Li, Xin and Wen, Chenglu and Wang, Cheng},
  booktitle={CVPR},
  pages={15321--15330},
  year={2024}
}

@inproceedings{boost_modal,
  title = {Boosting {{3D Object Detection}} by {{Simulating Multimodality}} on {{Point Clouds}}},
  booktitle = {{{CVPR}}},
  author = {Zheng, Wu and Hong, Mingxuan and Jiang, Li and Fu, Chi-Wing},
  year = {2022}
}

@inproceedings{boost_frame,
  title = {Boosting {{Single-Frame 3D Object Detection}} by {{Simulating Multi-Frame Point Clouds}}},
  booktitle = {ACMMM},
  author = {Zheng, Wu and Jiang, Li and Lu, Fanbin and Ye, Yangyang and Fu, Chi-Wing},
  year = {2022},
}

@inproceedings{pointrcnn,
  title = {{{PointRCNN}}: {{3D Object Proposal Generation}} and {{Detection From Point Cloud}}},
  booktitle = {{{CVPR}}},
  author = {Shi, Shaoshuai and Wang, Xiaogang and Li, Hongsheng},
  year = {2019}
}

@inproceedings{pvrcnn,
  title = {{{PV-RCNN}}: {{Point-Voxel Feature Set Abstraction}} for {{3D Object Detection}}},
  booktitle = {{{CVPR}}},
  author = {Shi, Shaoshuai and Guo, Chaoxu and Jiang, Li and Wang, Zhe and Shi, Jianping and Wang, Xiaogang and Li, Hongsheng},
  year = {2020}
}

@article{second,
  title = {{{SECOND}}: {{Sparsely Embedded Convolutional Detection}}},
  author = {Yan, Yan and Mao, Yuxing and Li, Bo},
  year = {2018},
  journal = {Sensors},
  volume = {18}
}

@inproceedings{virconv,
  title = {Virtual {{Sparse Convolution}} for {{Multimodal 3D Object Detection}}},
  booktitle = {{{CVPR}}},
  author = {Wu, Hai and Wen, Chenglu and Shi, Shaoshuai and Li, Xin and Wang, Cheng},
  year = {2023}
}

@inproceedings{cbam,
  title={Cbam: Convolutional block attention module},
  author={Woo, Sanghyun and Park, Jongchan and Lee, Joon-Young and Kweon, In So},
  booktitle={Proceedings of the European conference on computer vision (ECCV)},
  pages={3--19},
  year={2018}
}

@article{voxelrcnn,
  title = {Voxel {{R-CNN}}: {{Towards High Performance Voxel-based 3D Object Detection}}},
  author = {Deng, Jiajun and Shi, Shaoshuai and Li, Peiwei and Zhou, Wengang and Zhang, Yanyong and Li, Houqiang},
  year = {2021},
  journal = {AAAI},
  volume = {35}
}

@InProceedings{CenterPoint,
    author    = {Yin, Tianwei and Zhou, Xingyi and Krahenbuhl, Philipp},
    title     = {Center-Based 3D Object Detection and Tracking},
    booktitle = {Proceedings of the IEEE/CVF Conference on Computer Vision and Pattern Recognition (CVPR)},
    month     = {June},
    year      = {2021},
    pages     = {11784-11793}
}

@inproceedings{SRKD,
  title={Sunshine to rainstorm: Cross-weather knowledge distillation for robust 3d object detection},
  author={Huang, Xun and Wu, Hai and Li, Xin and Fan, Xiaoliang and Wen, Chenglu and Wang, Cheng},
  booktitle={Proceedings of the AAAI Conference on Artificial Intelligence},
  volume={38},
  pages={2409--2416},
  year={2024}
}

@InProceedings{coin,
    author    = {Xia, Qiming and Deng, Jinhao and Wen, Chenglu and Wu, Hai and Shi, Shaoshuai and Li, Xin and Wang, Cheng},
    title     = {CoIn: Contrastive Instance Feature Mining for Outdoor 3D Object Detection with Very Limited Annotations},
    booktitle = {Proceedings of the IEEE/CVF International Conference on Computer Vision (ICCV)},
    month     = {October},
    year      = {2023},
    pages     = {6254-6263}
}

@inproceedings{Dota,
  title={Learning to Detect Objects from Multi-Agent LiDAR Scans without Manual Labels},
  author={Xia, Qiming and Lin, Wenkai and Xiang, Haoen and Huang, Xun and Chen, Siheng and Dong, Zhen and Wang, Cheng and Wen, Chenglu},
  booktitle={Proceedings of the Computer Vision and Pattern Recognition Conference},
  pages={1418--1428},
  year={2025}
}

@article{survey_3DOD_robust_lidar,
  author={Song, Ziying and Liu, Lin and Jia, Feiyang and Luo, Yadan and Jia, Caiyan and Zhang, Guoxin and Yang, Lei and Wang, Li},
  journal={IEEE Transactions on Intelligent Transportation Systems}, 
  title={Robustness-Aware 3D Object Detection in Autonomous Driving: A Review and Outlook}, 
  year={2024},
  volume={25},
  number={11},
  pages={15407-15436},
  doi={10.1109/TITS.2024.3439557}
}

@article{survey_ijcv23,
  title={3D object detection for autonomous driving: A comprehensive survey},
  author={Mao, Jiageng and Shi, Shaoshuai and Wang, Xiaogang and Li, Hongsheng},
  journal={International Journal of Computer Vision},
  volume={131},
  number={8},
  pages={1909--1963},
  year={2023},
  publisher={Springer}
}

@InProceedings{Towards2024,
    author    = {Chae, Yujeong and Kim, Hyeonseong and Yoon, Kuk-Jin},
    title     = {Towards Robust 3D Object Detection with LiDAR and 4D Radar Fusion in Various Weather Conditions},
    booktitle = {Proceedings of the IEEE/CVF Conference on Computer Vision and Pattern Recognition (CVPR)},
    month     = {June},
    year      = {2024},
    pages     = {15162-15172}
}

@article{Lidar_for_autonomous_driving,
  title={Lidar for autonomous driving: The principles, challenges, and trends for automotive lidar and perception systems},
  author={Li, You and Ibanez-Guzman, Javier},
  journal={IEEE Signal Processing Magazine},
  volume={37},
  number={4},
  pages={50--61},
  year={2020},
  publisher={IEEE}
}

@article{occluded_objects_detection,
  title={A review of occluded objects detection in real complex scenarios for autonomous driving},
  author={Ruan, Jiageng and Cui, Hanghang and Huang, Yuhan and Li, Tongyang and Wu, Changcheng and Zhang, Kaixuan},
  journal={Green energy and intelligent transportation},
  volume={2},
  number={3},
  pages={100092},
  year={2023},
  publisher={Elsevier}
}

@article{survey_Towards_V2X,
  title={Towards vehicle-to-everything autonomous driving: A survey on collaborative perception},
  author={Liu, Si and Gao, Chen and Chen, Yuan and Peng, Xingyu and Kong, Xianghao and Wang, Kun and Xu, Runsheng and Jiang, Wentao and Xiang, Hao and Ma, Jiaqi and others},
  journal={arXiv preprint arXiv:2308.16714},
  year={2023}
}

@article{coalign,
  title={Robust collaborative 3d object detection in presence of pose errors},
  author={Lu, Yifan and Li, Quanhao and Liu, Baoan and Dianati, Mehrdad and Feng, Chen and Chen, Siheng and Wang, Yanfeng},
  journal={arXiv preprint arXiv:2211.07214},
  year={2022}
}

@inproceedings{cosdh,
  title={CoSDH: Communication-Efficient Collaborative Perception via Supply-Demand Awareness and Intermediate-Late Hybridization},
  author={Xu, Junhao and Zhang, Yanan and Cai, Zhi and Huang, Di},
  booktitle={Proceedings of the Computer Vision and Pattern Recognition Conference},
  pages={6834--6843},
  year={2025}
}

@article{bm2cp,
  title={Bm2cp: Efficient collaborative perception with lidar-camera modalities},
  author={Zhao, Binyu and Zhang, Wei and Zou, Zhaonian},
  journal={arXiv preprint arXiv:2310.14702},
  year={2023}
}

@article{survey_CO_Methods_datasets_challenges,
  title={Collaborative perception in autonomous driving: Methods, datasets, and challenges},
  author={Han, Yushan and Zhang, Hui and Li, Huifang and Jin, Yi and Lang, Congyan and Li, Yidong},
  journal={IEEE Intelligent Transportation Systems Magazine},
  volume={15},
  number={6},
  pages={131--151},
  year={2023},
  publisher={IEEE}
}

@inproceedings{attfuse,
  title={Opv2v: An open benchmark dataset and fusion pipeline for perception with vehicle-to-vehicle communication},
  author={Xu, Runsheng and Xiang, Hao and Xia, Xin and Han, Xu and Li, Jinlong and Ma, Jiaqi},
  booktitle={2022 International Conference on Robotics and Automation (ICRA)},
  pages={2583--2589},
  year={2022},
  organization={IEEE}
}

@inproceedings{V2X-ViT,
  title={V2X-ViT: Vehicle-to-everything cooperative perception with vision transformer},
  author={Xu, Runsheng and Xiang, Hao and Tu, Zhengzhong and Xia, Xin and Yang, Ming-Hsuan and Ma, Jiaqi},
  booktitle={European conference on computer vision},
  pages={107--124},
  year={2022},
  organization={Springer}
}

@inproceedings{adafusion,
  title={Adaptive feature fusion for cooperative perception using lidar point clouds},
  author={Qiao, Donghao and Zulkernine, Farhana},
  booktitle={Proceedings of the IEEE/CVF winter conference on applications of computer vision},
  pages={1186--1195},
  year={2023}
}

@article{PillarNeSt,
  author={Mao, Weixin and Wang, Tiancai and Zhang, Diankun and Yan, Junjie and Yoshie, Osamu},
  journal={IEEE Transactions on Intelligent Vehicles}, 
  title={PillarNeSt: Embracing Backbone Scaling and Pretraining for Pillar-based 3D Object Detection}, 
  year={2024},
  volume={},
  number={},
  pages={1-10},
  doi={10.1109/TIV.2024.3386576}}

@article{survey_co_IV,
  title={A survey of collaborative perception in intelligent vehicles at intersections},
  author={Gao, Xin and Zhang, Xinyu and Lu, Yiguo and Huang, Yuning and Yang, Lei and Xiong, Yijin and Liu, Peng},
  journal={IEEE Transactions on Intelligent Vehicles},
  year={2024},
  publisher={IEEE}
}

@inproceedings{point-gnn,
  title={Point-gnn: Graph neural network for 3d object detection in a point cloud},
  author={Shi, Weijing and Rajkumar, Raj},
  booktitle={Proceedings of the IEEE/CVF conference on computer vision and pattern recognition},
  pages={1711--1719},
  year={2020}
}

@article{disconet,
  title={Learning distilled collaboration graph for multi-agent perception},
  author={Li, Yiming and Ren, Shunli and Wu, Pengxiang and Chen, Siheng and Feng, Chen and Zhang, Wenjun},
  journal={Advances in Neural Information Processing Systems},
  volume={34},
  pages={29541--29552},
  year={2021}
}

@inproceedings{cooper,
  title={Cooper: Cooperative perception for connected autonomous vehicles based on 3d point clouds},
  author={Chen, Qi and Tang, Sihai and Yang, Qing and Fu, Song},
  booktitle={2019 IEEE 39th International Conference on Distributed Computing Systems (ICDCS)},
  pages={514--524},
  year={2019},
  organization={IEEE}
}

@article{Double-M_Quantification,
  title={Uncertainty quantification of collaborative detection for self-driving},
  author={Su, Sanbao and Li, Yiming and He, Sihong and Han, Songyang and Feng, Chen and Ding, Caiwen and Miao, Fei},
  journal={arXiv preprint arXiv:2209.08162},
  year={2022}
}

@inproceedings{mash,
  title={Overcoming obstructions via bandwidth-limited multi-agent spatial handshaking},
  author={Glaser, Nathaniel and Liu, Yen-Cheng and Tian, Junjiao and Kira, Zsolt},
  booktitle={2021 IEEE/RSJ International Conference on Intelligent Robots and Systems (IROS)},
  pages={2406--2413},
  year={2021},
  organization={IEEE}
}

@article{mamp,
  title={Model-agnostic multi-agent perception framework},
  author={Xu, Runsheng and Chen, Weizhe and Xiang, Hao and Liu, Lantao and Ma, Jiaqi},
  journal={arXiv preprint arXiv:2203.13168},
  year={2022}
}

@inproceedings{Di-v2x,
  title={Di-v2x: Learning domain-invariant representation for vehicle-infrastructure collaborative 3d object detection},
  author={Li, Xiang and Yin, Junbo and Li, Wei and Xu, Chengzhong and Yang, Ruigang and Shen, Jianbing},
  booktitle={Proceedings of the AAAI Conference on Artificial Intelligence},
  volume={38},
  pages={3208--3215},
  year={2024}
}

@article{MKD-cooper,
  title={MKD-cooper: Cooperative 3D object detection for autonomous driving via multi-teacher knowledge distillation},
  author={Li, Zhiyuan and Liang, Huawei and Wang, Hanqi and Zhao, Mingzhuo and Wang, Jian and Zheng, Xiaokun},
  journal={IEEE Transactions on Intelligent Vehicles},
  volume={9},
  number={1},
  pages={1490--1500},
  year={2023},
  publisher={IEEE}
}

@article{Scnet3d,
  title={Scnet3d: Rethinking the feature extraction process of pillar-based 3d object detection},
  author={Li, Junru and Wang, Zhiling and Gong, Diancheng and Wang, Chunchun},
  journal={IEEE Transactions on Intelligent Transportation Systems},
  year={2024},
  publisher={IEEE}
}

@article{Sparse2Dense,
  title={Sparse2Dense: Learning to densify 3d features for 3d object detection},
  author={Wang, Tianyu and Hu, Xiaowei and Liu, Zhengzhe and Fu, Chi-Wing},
  journal={Advances in Neural Information Processing Systems},
  volume={35},
  pages={38533--38545},
  year={2022}
}

@inproceedings{v2v4real,
  title={V2v4real: A real-world large-scale dataset for vehicle-to-vehicle cooperative perception},
  author={Xu, Runsheng and Xia, Xin and Li, Jinlong and Li, Hanzhao and Zhang, Shuo and Tu, Zhengzhong and Meng, Zonglin and Xiang, Hao and Dong, Xiaoyu and Song, Rui and others},
  booktitle={Proceedings of the IEEE/CVF conference on computer vision and pattern recognition},
  pages={13712--13722},
  year={2023}
}

@inproceedings{2025aaaidsrc,
  title={Dsrc: Learning density-insensitive and semantic-aware collaborative representation against corruptions},
  author={Zhang, Jingyu and Wang, Yilei and Qian, Lang and Sun, Peng and Li, Zengwen and Jiang, Sudong and Liu, Maolin and Song, Liang},
  booktitle={Proceedings of the AAAI Conference on Artificial Intelligence},
  volume={39},
  pages={9942--9950},
  year={2025}
}

@article{r2ldm,
  title={R2LDM: An Efficient 4D Radar Super-Resolution Framework Leveraging Diffusion Model},
  author={Zheng, Boyuan and Lu, Shouyi and Huang, Renbo and Huang, Minqing and Lu, Fan and Tian, Wei and Zhuo, Guirong and Xiong, Lu},
  journal={arXiv preprint arXiv:2503.17097},
  year={2025}
}

@article{where2comm,
  title={Where2comm: Communication-efficient collaborative perception via spatial confidence maps},
  author={Hu, Yue and Fang, Shaoheng and Lei, Zixing and Zhong, Yiqi and Chen, Siheng},
  journal={Advances in neural information processing systems},
  volume={35},
  pages={4874--4886},
  year={2022}
}

@inProceedings{V2X-R_CVPR,
    author    = {Huang, Xun and Wang, Jinlong and Xia, Qiming and Chen, Siheng and Yang, Bisheng and Li, Xin and Wang, Cheng and Wen, Chenglu},
    title     = {V2X-R: Cooperative LiDAR-4D Radar Fusion with Denoising Diffusion for 3D Object Detection},
    booktitle = {Proceedings of the IEEE/CVF Conference on Computer Vision and Pattern Recognition (CVPR)},
    month     = {June},
    year      = {2025},
    pages     = {27390-27400}
}

@inproceedings{dair-v2x,
  title={Dair-v2x: A large-scale dataset for vehicle-infrastructure cooperative 3d object detection},
  author={Yu, Haibao and Luo, Yizhen and Shu, Mao and Huo, Yiyi and Yang, Zebang and Shi, Yifeng and Guo, Zhenglong and Li, Hanyu and Hu, Xing and Yuan, Jirui and Nie, Zaiqing},
  booktitle={Proceedings of the IEEE/CVF Conference on Computer Vision and Pattern Recognition},
  pages={21361--21370},
  year={2022}
}

@inproceedings{cirkd,
  title={Cross-image relational knowledge distillation for semantic segmentation},
  author={Yang, Chuanguang and Zhou, Helong and An, Zhulin and Jiang, Xue and Xu, Yongjun and Zhang, Qian},
  booktitle={Proceedings of the IEEE/CVF conference on computer vision and pattern recognition},
  pages={12319--12328},
  year={2022}
}

@inproceedings{1-supfusion,
  title={SupFusion: Supervised LiDAR-camera fusion for 3D object detection},
  author={Qin, Yiran and Wang, Chaoqun and Kang, Zijian and Ma, Ningning and Li, Zhen and Zhang, Ruimao},
  booktitle={Proceedings of the IEEE/CVF international conference on computer vision},
  pages={22014--22024},
  year={2023}
}

@inproceedings{2-2dpass,
  title={2dpass: 2d priors assisted semantic segmentation on lidar point clouds},
  author={Yan, Xu and Gao, Jiantao and Zheng, Chaoda and Zheng, Chao and Zhang, Ruimao and Cui, Shuguang and Li, Zhen},
  booktitle={European conference on computer vision},
  pages={677--695},
  year={2022},
  organization={Springer}
}

@article{3-learning,
  title={Learning 3D Perception from Others' Predictions},
  author={Yoo, Jinsu and Feng, Zhenyang and Pan, Tai-Yu and Sun, Yihong and Phoo, Cheng Perng and Chen, Xiangyu and Campbell, Mark and Weinberger, Kilian Q and Hariharan, Bharath and Chao, Wei-Lun},
  journal={arXiv preprint arXiv:2410.02646},
  year={2024}
}

@article{5-V2X-INCOP,
  title={Interruption-aware cooperative perception for V2X communication-aided autonomous driving},
  author={Ren, Shunli and Lei, Zixing and Wang, Zi and Dianati, Mehrdad and Wang, Yafei and Chen, Siheng and Zhang, Wenjun},
  journal={IEEE Transactions on Intelligent Vehicles},
  volume={9},
  number={4},
  pages={4698--4714},
  year={2024},
  publisher={IEEE}
}
\end{document}